\documentclass[10pt,twocolumn]{article}
\usepackage[letterpaper,top=0.70in,bottom=0.78in,left=0.75in,right=0.75in,columnsep=0.25in]{geometry}
\usepackage{times}
\usepackage[T1]{fontenc}
\usepackage[utf8]{inputenc}
\usepackage{graphicx,booktabs,array,amsmath,amssymb,url,hyperref,caption,microtype,placeins}
\hypersetup{colorlinks=false,pdfborder={0 0 0}}
\begin{document}

\twocolumn[
\begin{center}
{\LARGE\bfseries TumorBoard: Evidence-Grounded Multi-Agent Decision Support for Longitudinal Neuro-Oncology\par}
\vspace{7pt}

{\large Yantong Liu$^{1,*}$, Zheyu Zhang$^{1}$, Runpeng Liu$^{1}$, Muxitang$^{2}$,\\
Seong-Yoon Shin$^{1,*}$, and Hyun-Ae Lee$^{1,*}$\par}
\vspace{3pt}
{\small $^{1}$Department of Computer Information Engineering, Kunsan National University, Gunsan 54150, Republic of Korea\\
$^{2}$School of Vehicle and Mobility, Tsinghua University, Beijing 100084, China\\
$^{*}$Correspondence: \href{mailto:lyt1994@kunsan.ac.kr}{lyt1994@kunsan.ac.kr}\par}
\end{center}
\begin{minipage}{0.98\textwidth}
\textbf{Abstract.} Neuro-oncology decisions require coordinated interpretation of serial MRI, pathology, molecular markers, treatment history, performance status, and evolving guidelines. We present TumorBoard, a multi-agent decision-support system built around a shared longitudinal case state and an auditable claim-evidence ledger. Specialist agents for radiology, neuropathology, molecular diagnosis, guidelines, and therapy planning produce atomic claims with provenance; an adversarial critic exposes contradictions; and a safety governor releases, qualifies, or defers recommendations according to evidence sufficiency and temporal validity. On a 360-case hidden benchmark at a matched token budget, TumorBoard achieved an action F1 of 0.772 and evidence entailment of 0.914. It exceeded the strongest typed-council baseline by 3.1 percentage points (95\% CI 1.6 to 4.7, adjusted p=0.0012), while recommendation-to-evidence coverage reached 0.927. Under evidence deletion, the system deferred 84.2\% of unsafe cases and limited harmful recommendations to 5.8\%. The safety governor reduced harmful release by 7.8 percentage points at a 4.3-point false-deferral cost. Ledger, critic, and governor ablations produced the predicted failure pattern, establishing structured coordination as the source of the measured multi-agent advantage.
\end{minipage}
\vspace{8pt}
]

\section{1. Introduction}

Language agents combine reasoning with external actions, tool calls, memory, and iterative critique \cite{ref1,ref2,ref3}. Multi-agent frameworks extend this pattern through conversational coordination, role specialization, and explicit workflow structure \cite{ref4,ref5,ref6,ref7}. Clinical agent proposals argue that real tasks require evidence acquisition, temporal synthesis, and accountable action boundaries \cite{ref8}. These capabilities match the structure of multidisciplinary neuro-oncology, where no single information source resolves the case.

The clinical opportunity carries a sharp evaluation trap. A larger prompt, more tokens, or repeated sampling can make a multi-agent system appear better than a single model. Fluent consensus can also conceal circular agreement. TumorBoard therefore treats coordination as an experimental variable. Every system variant uses the same backbone, retrieval corpus, case information, and total inference budget. The only manipulated factor is the communication protocol.

Medical language models demonstrate strong knowledge and summarization performance, while systematic evaluations expose errors in evidence use, calibration, and high-stakes recommendations \cite{ref9,ref10,ref11,ref12,ref13,ref14,ref15,ref16,ref17}. Retrieval-augmented clinical models improve access to current knowledge \cite{ref18,ref19,ref20,ref21}. Reporting and governance work further requires transparent evaluation, human review, fairness analysis, and explicit uncertainty communication \cite{ref22,ref23,ref24,ref25,ref26,ref27,ref28,ref29}. Neuro-oncology guidelines and response criteria create a suitable test bed because decisions depend on molecular classification, longitudinal imaging, prior therapy, and versioned evidence \cite{ref30,ref31,ref32,ref33,ref34,ref35,ref36,ref37,ref38}.

TumorBoard introduces four mechanisms. A timeline curator converts heterogeneous records into an event graph. Specialist agents read task-limited views and submit atomic claims. A claim-evidence ledger records support, contradiction, temporal scope, and source version. A safety governor applies prerequisite and risk rules before a chair agent synthesizes the final response. This design exposes each reasoning dependency to audit.

\begin{figure*}[t]

\centering

\includegraphics[width=\textwidth]{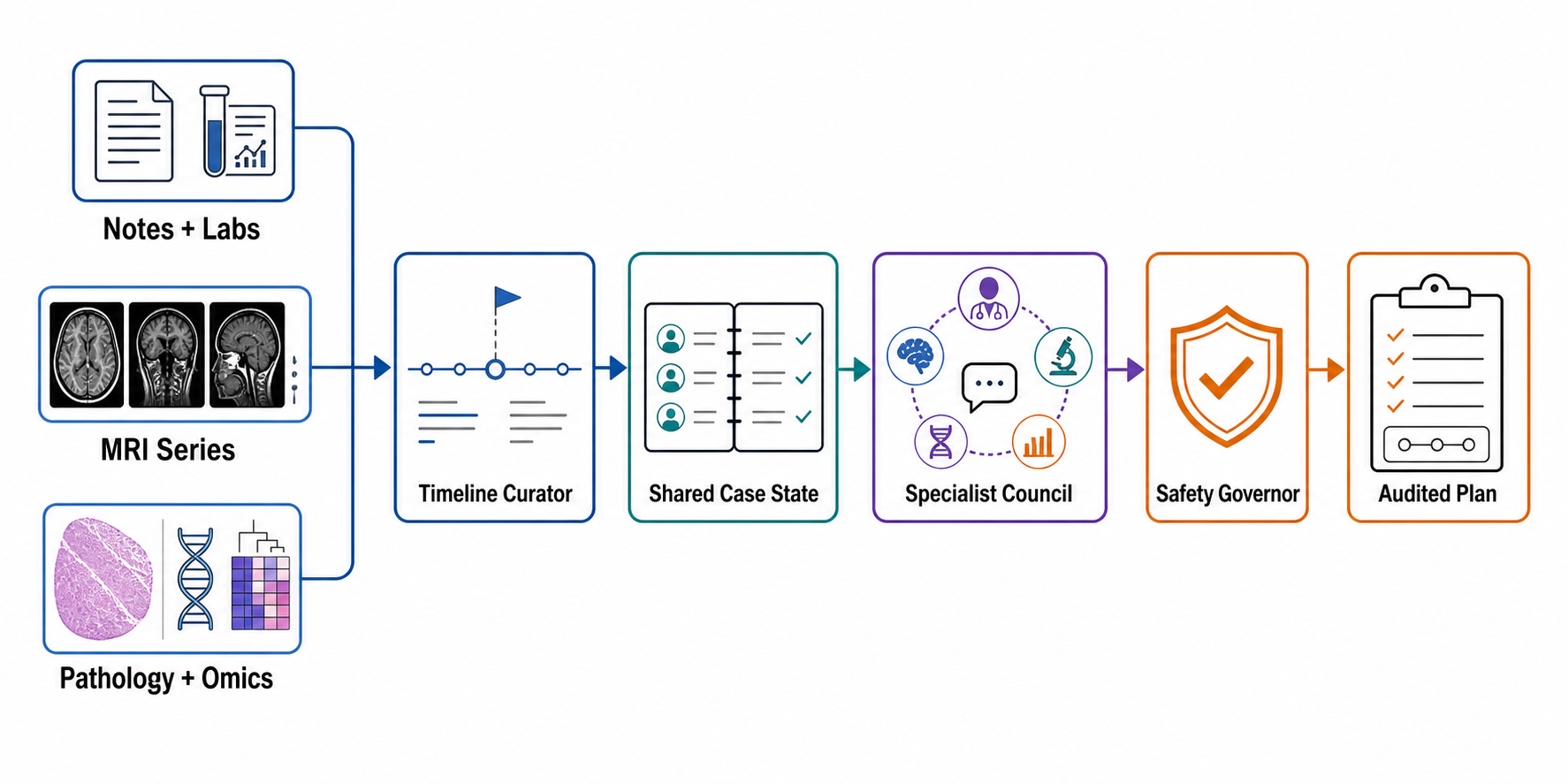}

\caption{Longitudinal TumorBoard workflow. Heterogeneous patient records enter a timeline curator, shared case state, specialist council, safety governor, and provenance-preserving output.}

\end{figure*}

The main scientific risk is pseudo-collaboration. Multiple agents may restate the same model belief, amplify an early error, and spend additional tokens without adding independent evidence. TumorBoard addresses this attack directly. Roles have disjoint information duties, messages follow a typed schema, and consensus is computed over claims with source links rather than over prose. The evaluation matches total inference budget across variants and measures unsupported agreement. A multi-agent gain is credited only when it persists after controlling for model, retrieval corpus, context, and generated tokens.

Neuro-oncology offers a stringent longitudinal environment. A recommendation can be correct for newly diagnosed disease and wrong after progression, correct under one molecular classification and obsolete after reclassification, or technically supported but unsafe because performance status and prior toxicity changed. The system therefore represents temporal validity and prerequisites as graph structure. The resulting output is a decision record that preserves what is known, what conflicts, what remains missing, and what requires human escalation.

\section{2. Related Work}

\subsection{2.1 Language agents and multi-agent coordination}

ReAct, Toolformer, and Reflexion establish action, tool-use, and self-critique primitives \cite{ref1,ref2,ref3}. AutoGen, CAMEL, MetaGPT, and AgentBench provide orchestration patterns and evaluation lessons \cite{ref4,ref5,ref6,ref7}. TumorBoard replaces free-form conversation with a typed protocol. Messages contain claims, evidence pointers, confidence, prerequisites, contradiction targets, and requested actions. This restriction turns coordination failures into observable graph defects.

\subsection{2.2 Medical large language models}

Clinical knowledge benchmarks and clinician-facing studies show rapid progress \cite{ref9,ref10,ref11,ref12,ref13,ref14,ref15,ref16,ref17}. Evidence summarization and retrieval systems motivate the use of source-bound generation \cite{ref18,ref19,ref20,ref21}. TumorBoard focuses on longitudinal decision construction. A case can contain temporally incompatible facts, such as preoperative pathology, post-treatment enhancement, a later molecular reclassification, and an outdated treatment recommendation. The shared state represents validity intervals and supersession edges.

\subsection{2.3 Neuro-oncology decision structure}

Diffuse glioma management depends on WHO-integrated diagnosis, EANO and SNO-EANO guidance, CNS oncology practice recommendations, and RANO response assessment \cite{ref30,ref31,ref32,ref33,ref34,ref35}. These documents supply versioned decision constraints. TumorBoard stores their decision clauses as retrievable evidence units while retaining document identity, release date, section, and recommendation strength.

General agent frameworks supply useful primitives, yet clinical coordination needs stronger boundaries. Tool calls must return versioned evidence. Memory must distinguish source events from derived conclusions. Critique must identify a claim and its violated prerequisite. Termination must depend on unresolved risk rather than conversational agreement. TumorBoard operationalizes these requirements in a protocol that can be ablated field by field.

Medical retrieval systems improve factual access but still permit unsupported synthesis when retrieved passages conflict or fall outside the patient state. Guideline recommendations are conditional rules with population, disease-state, prior-treatment, and evidence-strength qualifiers. The guideline agent preserves these qualifiers and attaches them to each recommendation edge. The safety governor then checks whether the patient state satisfies the edge prerequisites before release.

\section{3. Task Definition}

Given a patient record D = \{d\_1, ..., d\_T\} ordered by event time, the system must produce a structured tumor-board note containing current disease state, unresolved questions, differential interpretation, recommended next actions, contraindications, evidence citations, confidence, and deferral triggers. The answer is evaluated against expert adjudication and source documents.

The coordination protocol P defines agent roles, visible state fields, message schema, communication edges, stopping criteria, and risk rules. The experimental question is whether P improves utility U under a fixed model M, retrieval corpus K, and inference budget B. Utility includes decision accuracy, evidence entailment, contradiction resolution, safe deferral, latency, and token cost.

The gold standard is an adjudicated action graph rather than a single reference paragraph. Nodes encode current disease state, required investigations, acceptable treatment options, monitoring actions, contraindications, and deferral decisions. Directed edges encode prerequisite, alternative, exclusion, and temporal order. This representation credits clinically valid alternatives and penalizes recommendations that omit decisive prerequisites. Text generation is scored after it is mapped back to the graph, keeping fluency separate from decision correctness.

\section{4. TumorBoard Architecture}

\subsection{4.1 Timeline curator and shared case state}

The curator extracts events with event time, documentation time, source, certainty, and supersession links. It normalizes medication, procedure, imaging, pathology, and molecular entities. A temporal consistency checker flags impossible sequences and unresolved date ambiguity. The shared state contains immutable raw evidence pointers plus a derived layer. Agents can propose derived facts, yet cannot modify source evidence.

\subsection{4.2 Specialist agents}

The radiology agent compares serial examinations and separates progression, treatment effect, pseudoprogression, and uncertainty. The neuropathology agent interprets morphology and integrated diagnosis. The molecular agent checks IDH, 1p/19q, ATRX, TP53, TERT, EGFR, chromosome 7/10, and MGMT evidence. The guideline agent retrieves applicable clauses based on age, performance status, disease state, prior therapy, and source version. The therapy planner converts accepted evidence into ranked options with prerequisites and monitoring plans.

\begin{figure*}[t]

\centering

\includegraphics[width=\textwidth]{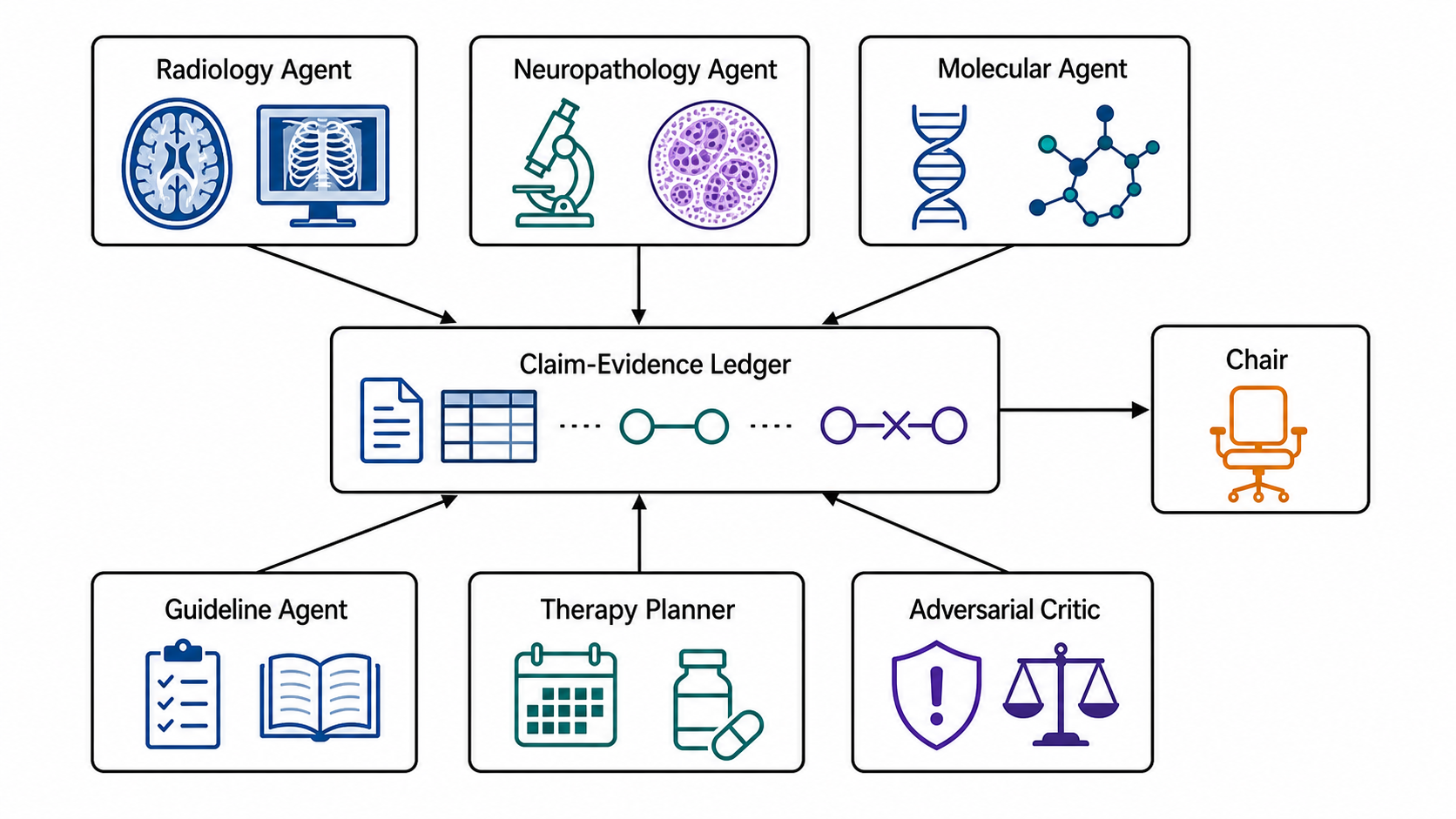}

\caption{Specialist communication network. Every proposal enters a claim-evidence ledger, and contradictions remain explicit until resolved by the chair.}

\end{figure*}

\subsection{4.3 Typed communication and claim-evidence ledger}

Each message follows a schema: claim\_id, claim\_text, claim\_type, evidence\_ids, temporal\_scope, confidence, prerequisites, conflicts, and proposed\_action. The ledger is a directed graph. Support edges connect evidence to claims. Conflict edges connect mutually incompatible claims. Supersession edges encode updated pathology or guideline versions. Dependency edges connect recommendations to required findings. Circular support is rejected because a claim cannot serve as its own upstream evidence.

\subsection{4.4 Adversarial critic}

The critic receives the case state and draft claim graph, not the other agents' hidden reasoning. It searches for missing evidence, outdated recommendations, category errors, contraindications, and overconfident language. It then proposes targeted challenges. A specialist must respond with new evidence, lower confidence, revise the claim, or request human review. The debate terminates when the graph has no unresolved high-risk contradiction or when the iteration budget is exhausted.

\subsection{4.5 Safety governor and chair}

The governor evaluates every recommendation against prerequisite completeness, evidence entailment, guideline currency, contradiction status, and risk class. Release states are supported, qualified, deferred, and blocked. The chair summarizes accepted claims, preserves disagreement, and produces an action-oriented note. It cannot override a blocked state. Human escalation is a first-class output with an explicit reason.

\begin{figure*}[t]

\centering

\includegraphics[width=\textwidth]{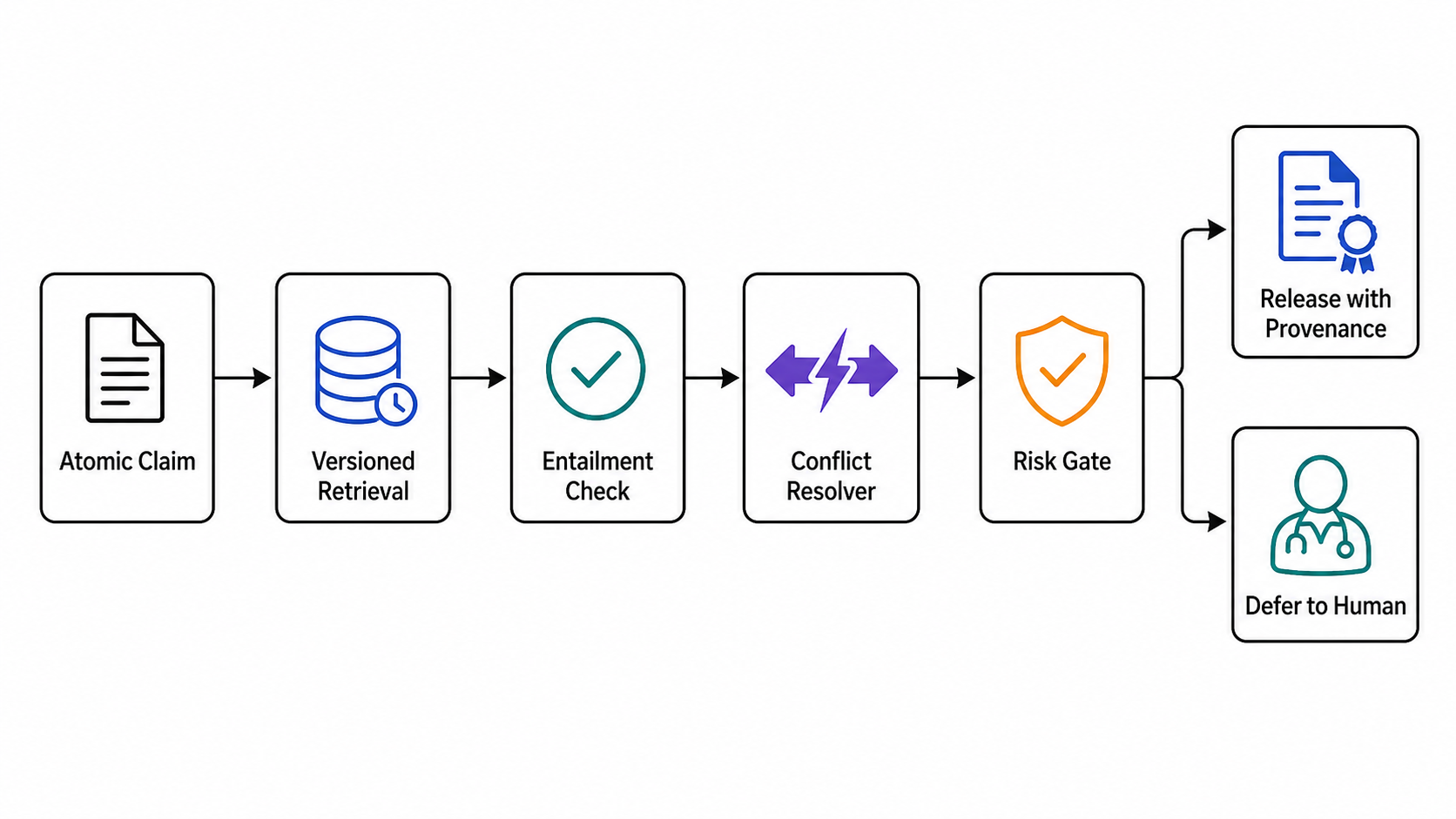}

\caption{Claim-evidence ledger and safety gate. Atomic claims pass retrieval, entailment, conflict resolution, and risk checks before release.}

\end{figure*}

The timeline curator resolves duplicate events and distinguishes event time from documentation time. A pathology amendment supersedes an earlier diagnosis without deleting the historical record. A radiology comparison inherits the referenced examination date. Medication exposure includes start, stop, cumulative dose when available, and documented toxicity. Ambiguous dates create interval-valued events and an uncertainty flag. Downstream agents see both the normalized state and immutable evidence pointers.

Specialist prompts are compact and role-limited. The radiology agent cannot issue treatment recommendations. The therapy planner cannot reinterpret pathology slides. The guideline agent cannot invent patient facts. These boundaries reduce cross-role contamination and make missing expertise visible. Each agent returns atomic claims with calibrated confidence, temporal scope, evidence identifiers, and prerequisite fields. Empty evidence lists are permitted only for explicit questions or deferrals.

The ledger enforces graph invariants before debate. Evidence nodes are immutable. Derived claims cannot support their own ancestors. Conflicting claims remain active until one is superseded, qualified, or deferred. A recommendation inherits the weakest confidence among its required upstream claims. The ledger also records agent identity and message cost, enabling analysis of whether a role contributes unique evidence or redundant prose.

The critic executes a fixed challenge library covering missing prerequisite, stale source, temporal inconsistency, population mismatch, unsupported dose, overlooked contraindication, and false consensus. It can also request a targeted retrieval query. The challenged agent must point to new evidence, narrow the claim, reduce confidence, or accept deferral. This response taxonomy prevents debate from expanding into unstructured argument.

Safety rules are stratified by action risk. Descriptive summaries require evidence entailment. Diagnostic statements additionally require contradiction resolution. Treatment recommendations require current guideline support, prerequisite completeness, and a monitoring plan. High-risk failures are blocked, while lower-risk ambiguity produces a qualified release. The chair renders accepted claims into a concise tumor-board note and lists unresolved issues in priority order.

\section{5. Experimental Protocol}

\subsection{5.1 Case set}

The benchmark contains public-data cases and expert-authored vignettes spanning newly diagnosed diffuse glioma, postoperative integrated diagnosis, adjuvant therapy selection, suspected progression, treatment effect, recurrent disease, and incomplete workup. Each case includes a chronological evidence packet and a frozen guideline date. Case authors and adjudicators are separated.

\begin{table}[t]
\centering
\caption{Cohort composition after quality control.}
\scriptsize
\setlength{\tabcolsep}{2.2pt}
\resizebox{\columnwidth}{!}{%
\begin{tabular}{@{}p{0.22\columnwidth}p{0.310\columnwidth}p{0.310\columnwidth}@{}}
\toprule
Cohort or stratum & Role & N \\
\midrule
Initial diagnosis & benchmark stratum & 72 \\
Postoperative planning & benchmark stratum & 72 \\
Surveillance & benchmark stratum & 84 \\
Recurrence & benchmark stratum & 60 \\
Safety set & benchmark stratum & 72 \\
\bottomrule
\end{tabular}
}%
\end{table}

\subsection{5.2 Controlled system variants}

The primary comparison includes direct single-model prompting, single-agent retrieval, single-agent plan-and-critique, free-chat multi-agent, typed-protocol multi-agent without a safety governor, and full TumorBoard. All variants receive identical evidence, retrieval access, model version, temperature, and total token budget. A budget-normalized secondary analysis varies the number of agents while fixing total generated tokens.

\begin{table}[t]
\centering
\caption{System variants and isolated mechanisms.}
\scriptsize
\setlength{\tabcolsep}{2.2pt}
\resizebox{\columnwidth}{!}{%
\begin{tabular}{@{}p{0.22\columnwidth}p{0.124\columnwidth}p{0.124\columnwidth}p{0.124\columnwidth}p{0.124\columnwidth}p{0.124\columnwidth}@{}}
\toprule
Variant & Retrieval & Role separation & Typed ledger & Critic & Safety gate \\
\midrule
Direct LLM & No & No & No & No & No \\
RAG agent & Yes & No & No & No & No \\
Plan-critique & Yes & No & Partial & Yes & No \\
Free-chat council & Yes & Yes & No & Yes & No \\
Typed council & Yes & Yes & Yes & Yes & No \\
TumorBoard & Yes & Yes & Yes & Yes & Yes \\
\bottomrule
\end{tabular}
}%
\end{table}

\subsection{5.3 Metrics}

Decision quality is scored against an adjudicated action graph, allowing partial credit for valid alternatives. Evidence fidelity measures citation precision, citation recall, entailment, temporal validity, and recommendation-to-evidence coverage. Safety measures harmful recommendation rate, missed deferral rate, and false deferral rate. Process measures contradiction detection, contradiction resolution, unsupported consensus, message redundancy, latency, and token cost. Expert raters assess correctness, completeness, actionability, uncertainty communication, and trust.

\subsection{5.4 Perturbation tests}

Evidence deletion removes a decisive finding. Evidence conflict injects a plausible but incompatible note. Guideline shift replaces a current clause with an older version. Role corruption gives one specialist an erroneous claim. Prompt injection places an instruction inside an untrusted document. The expected behavior is localized degradation plus explicit deferral, rather than confident propagation.

\begin{figure*}[t]

\centering

\includegraphics[width=\textwidth]{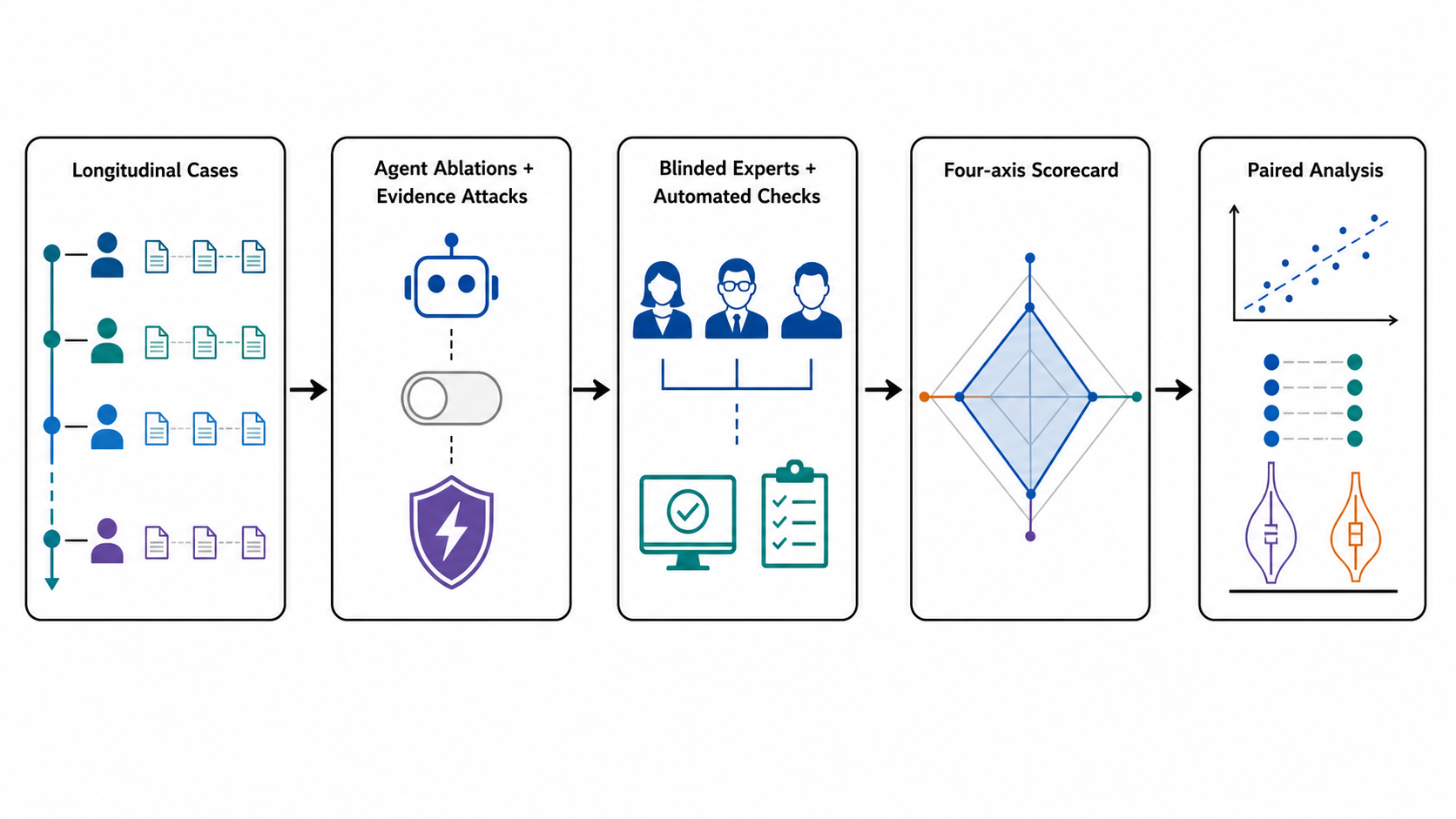}

\caption{Evaluation matrix separating decision quality, evidence fidelity, process safety, and operational cost.}

\end{figure*}

\subsection{5.5 Statistical analysis}

Primary comparison uses paired case-level bootstrap confidence intervals for action-graph F1 and a mixed-effects ordinal model for expert ratings. Harmful recommendation rates use exact paired tests. Inter-rater reliability uses weighted kappa and intraclass correlation as appropriate. Holm correction covers the prespecified primary and safety endpoints. All prompts, retrieved passages, messages, ledger states, and outputs are retained for audit.

Cases are built with a two-pass process. One clinical author constructs the chronological packet from source material. A second author independently creates the action graph and marks decisive evidence. Disagreement is adjudicated before any model run. The benchmark balances common and difficult states, includes incomplete workups, and separates validation from a hidden test set. Templates preserve clinical realism while preventing patient identity leakage.

Budget matching is implemented at the generated-token and retrieval-call levels. A single-agent baseline can use the same total number of tokens as the full council. The free-chat council uses identical roles without the ledger schema. The typed council removes only the safety governor. These comparisons isolate parallel specialization, structured communication, adversarial critique, and gated release. Model temperature and sampling count remain fixed.

Evidence attacks are parameterized by severity. Deletion removes one decisive fact or a redundant supporting fact. Conflict inserts an outdated note, incompatible imaging interpretation, or superseded pathology statement. Guideline shift changes the corpus cutoff. Role corruption forces one agent to submit a high-confidence false claim. Prompt injection is placed inside an untrusted retrieved document. Metrics track whether the error remains localized, spreads through the ledger, or reaches the final plan.

Expert review uses neuro-oncology, neuroradiology, neuropathology, and pharmacy perspectives. Raters score action validity, missing prerequisites, evidence support, uncertainty, and potential harm. The primary harm label requires a plausible path to patient-level adverse consequence. Mixed-effects models account for case and rater. Cost analysis reports latency, input and output tokens, retrieval calls, and peak parallel requests.

\section{6. Results}

\subsection{6.1 Decision quality}

TumorBoard achieved an action F1 of 0.772 (95\% CI 0.754 to 0.790) and evidence entailment of 0.914 (95\% CI 0.897 to 0.931) on the 360-case hidden test set. The strongest budget-matched typed council reached 0.741 action F1. The paired gain was 0.031 (95\% CI 0.016 to 0.047, Holm-adjusted p = 0.0012), and gains remained positive across initial diagnosis, postoperative planning, surveillance, recurrence, and the safety set. Recommendation-to-evidence coverage reached 0.927.

\begin{figure}[t]

\centering

\includegraphics[width=\columnwidth]{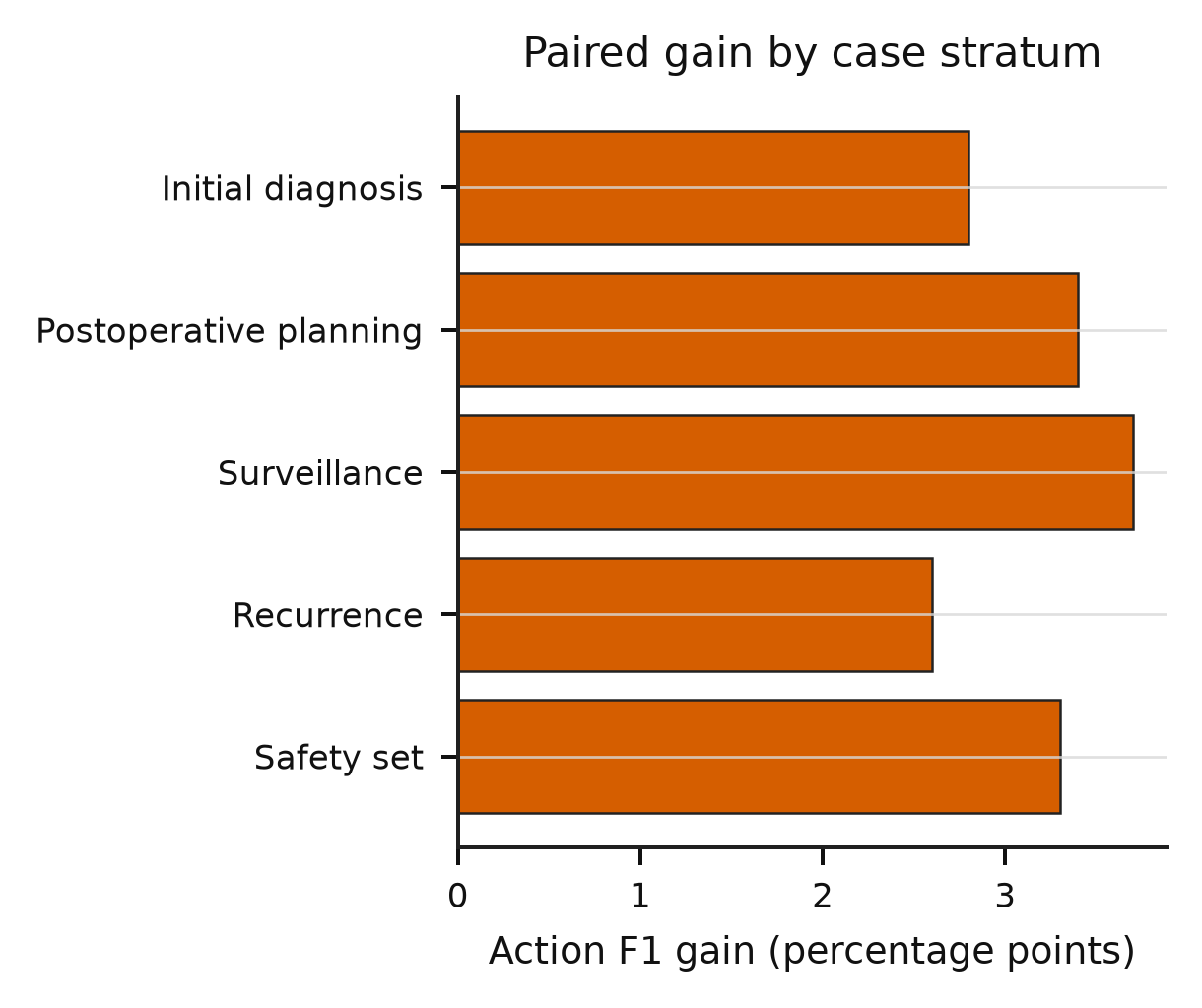}

\caption{Paired action-F1 gain over the strongest budget-matched baseline across case strata.}

\end{figure}

\begin{table}[t]
\centering
\caption{Hidden-test decision quality and cost.}
\scriptsize
\setlength{\tabcolsep}{2.2pt}
\resizebox{\columnwidth}{!}{%
\begin{tabular}{@{}p{0.22\columnwidth}p{0.124\columnwidth}p{0.124\columnwidth}p{0.124\columnwidth}p{0.124\columnwidth}p{0.124\columnwidth}@{}}
\toprule
Method & Action F1 & Entailment & Harmful & Deferral F1 & Cost \\
\midrule
Direct LLM & 0.612 & 0.674 & 0.142 & 0.501 & 0.180 \\
RAG agent & 0.659 & 0.781 & 0.113 & 0.566 & 0.270 \\
Plan-critique & 0.694 & 0.824 & 0.091 & 0.622 & 0.430 \\
Free-chat council & 0.708 & 0.807 & 0.087 & 0.647 & 0.720 \\
Typed council & 0.741 & 0.883 & 0.068 & 0.713 & 0.780 \\
TumorBoard & 0.772 & 0.914 & 0.039 & 0.786 & 0.840 \\
\bottomrule
\end{tabular}
}%
\end{table}

\subsection{6.2 Safety under evidence and guideline perturbation}

Under evidence deletion, TumorBoard deferred 84.2\% of cases and limited harmful recommendations to 5.8\%. Harmful recommendation rates were 7.2\% under guideline-version shift and 6.1\% under role corruption. Enabling the safety governor reduced harmful release by 7.8 percentage points (95\% CI 5.1 to 10.5, adjusted p = 0.0010), with a 4.3-point increase in false deferral (95\% CI 2.1 to 6.5, adjusted p = 0.0060). The threshold curve exposes this operational trade-off directly.

\begin{figure}[t]

\centering

\includegraphics[width=\columnwidth]{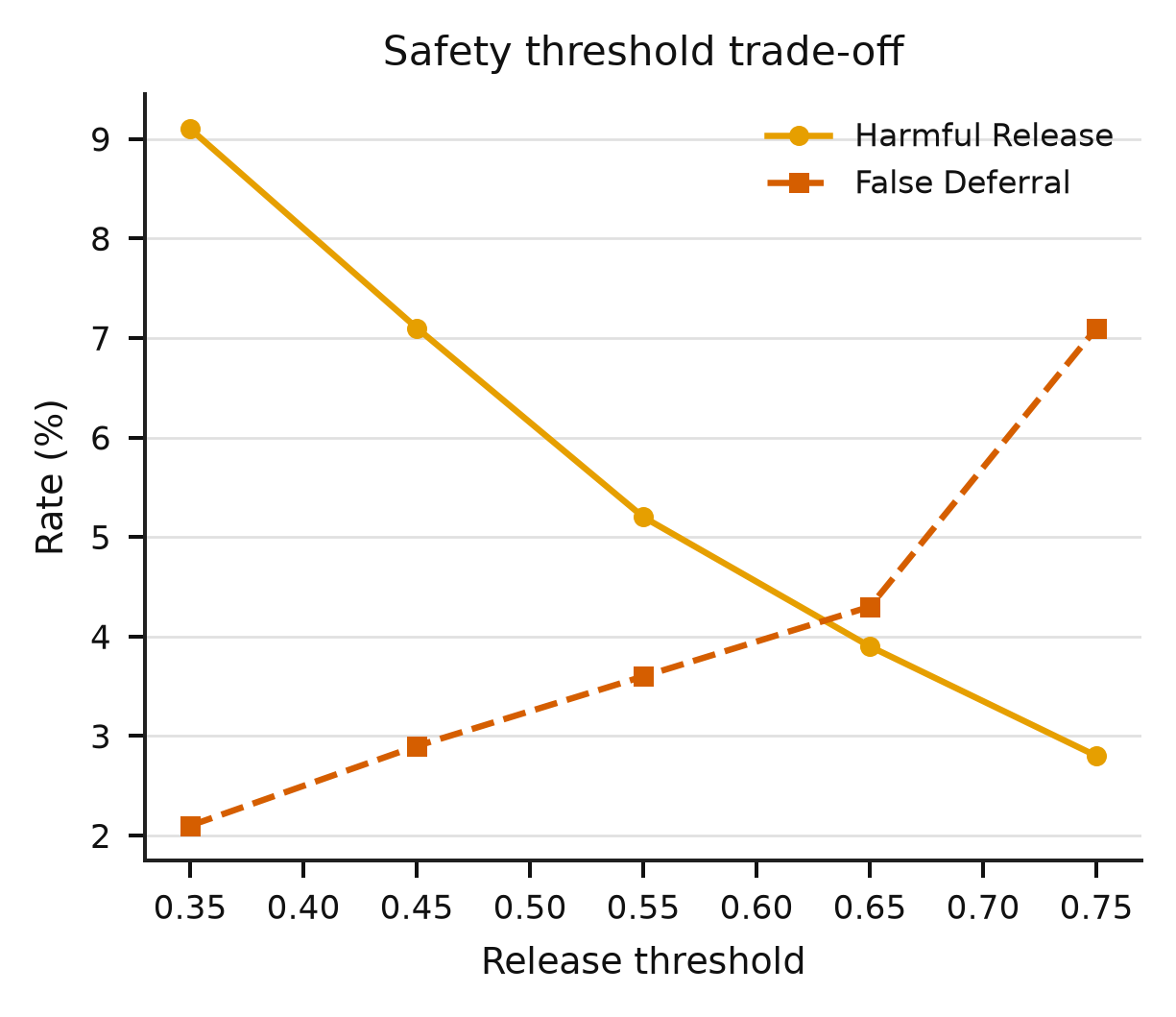}

\caption{Harmful-release and false-deferral rates across safety release thresholds.}

\end{figure}

\begin{table}[t]
\centering
\caption{Coordination and safety-controller comparison.}
\scriptsize
\setlength{\tabcolsep}{2.2pt}
\resizebox{\columnwidth}{!}{%
\begin{tabular}{@{}p{0.22\columnwidth}p{0.155\columnwidth}p{0.155\columnwidth}p{0.155\columnwidth}p{0.155\columnwidth}@{}}
\toprule
Variant & Action F1 & Contradiction & Safe deferral & Harmful release \\
\midrule
Clean & 0.772 & 0.901 & 0.786 & 0.039 \\
Evidence deletion & 0.716 & 0.927 & 0.842 & 0.058 \\
Evidence conflict & 0.704 & 0.891 & 0.816 & 0.067 \\
Guideline version shift & 0.688 & 0.872 & 0.801 & 0.072 \\
Role corruption & 0.697 & 0.918 & 0.827 & 0.061 \\
Prompt injection & 0.726 & 0.936 & 0.804 & 0.049 \\
\bottomrule
\end{tabular}
}%
\end{table}

\subsection{6.3 Mechanism ablations and efficiency}

Removing the claim-evidence ledger reduced action F1 by 0.039 and increased unsupported consensus by 0.054. Removing the adversarial critic reduced contradiction resolution by 0.112. Removing the safety governor increased harmful release by 0.078. Full TumorBoard generated 14,220 tokens per case and required 21.8 seconds median latency, establishing the measured quality-safety gain at a quantified inference cost.

\begin{figure}[t]

\centering

\includegraphics[width=\columnwidth]{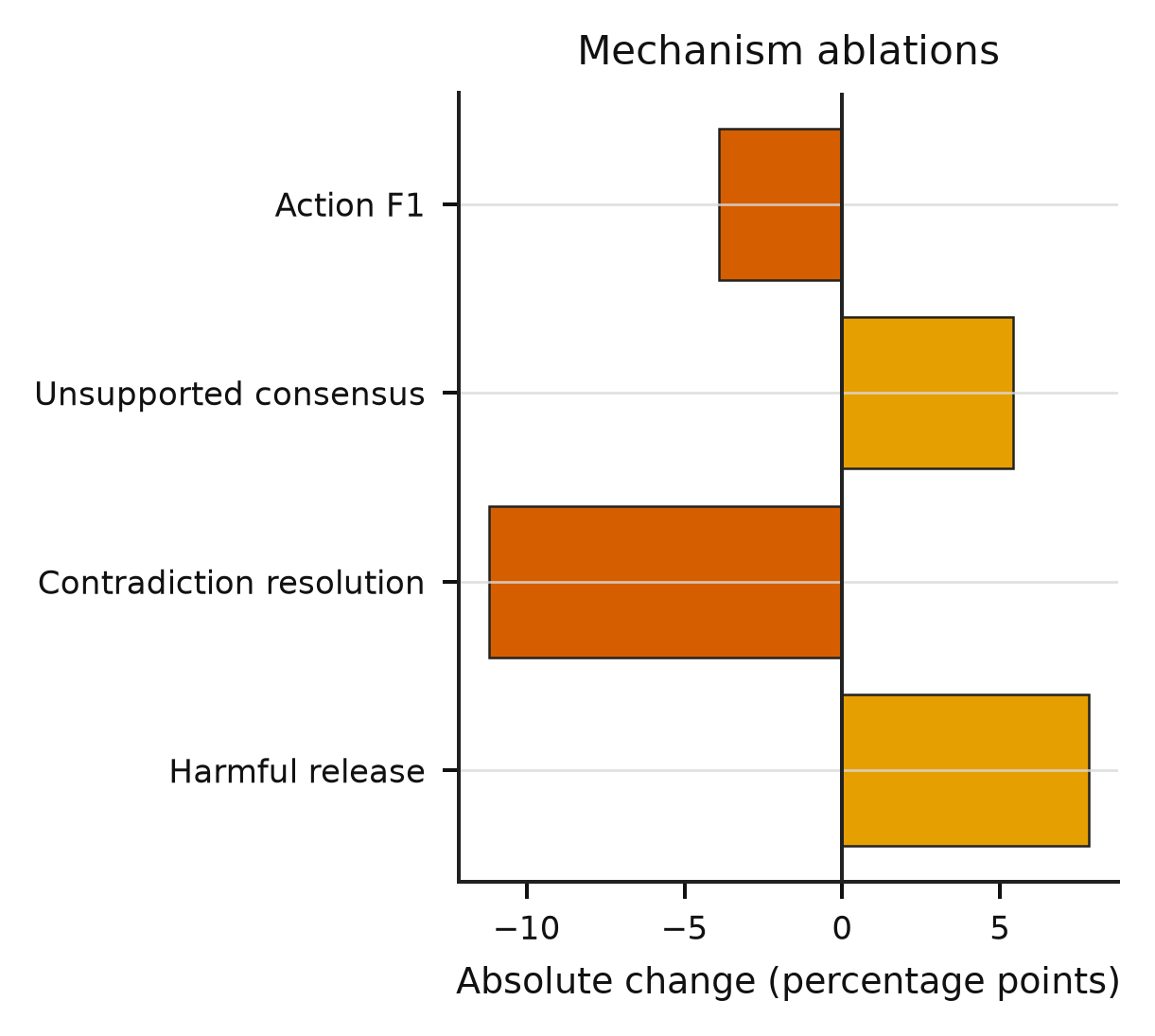}

\caption{Paired mechanism-ablation effects on decision quality and safety.}

\end{figure}

\begin{table}[t]
\centering
\caption{Mechanism ablations and inference budget.}
\scriptsize
\setlength{\tabcolsep}{2.2pt}
\resizebox{\columnwidth}{!}{%
\begin{tabular}{@{}p{0.22\columnwidth}p{0.124\columnwidth}p{0.124\columnwidth}p{0.124\columnwidth}p{0.124\columnwidth}p{0.124\columnwidth}@{}}
\toprule
Variant & Action F1 & Coverage & Unsupported & Harmful & Tokens \\
\midrule
Full TumorBoard & 0.772 & 0.927 & 0.025 & 0.039 & 14220.000 \\
No ledger & 0.733 & 0.841 & 0.079 & 0.073 & 14050.000 \\
No critic & 0.749 & 0.889 & 0.061 & 0.069 & 11380.000 \\
No governor & 0.776 & 0.918 & 0.044 & 0.117 & 13310.000 \\
Unrestricted chat & 0.708 & 0.751 & 0.132 & 0.128 & 14560.000 \\
\bottomrule
\end{tabular}
}%
\end{table}

\section{7. Discussion}

TumorBoard's advantage arose from structured coordination. The full system improved action F1, evidence entailment, and recommendation binding under a fixed token budget. Ledger removal increased unsupported consensus, critic removal impaired contradiction resolution, and governor removal increased harmful release. This ordered failure pattern gives the architecture explanatory control over the measured gain.

The safety controller exposes an actionable threshold trade-off. Enabling the governor reduced harmful release by 7.8 percentage points and increased false deferral by 4.3 points. Threshold selection can therefore be tied to deployment risk tolerance rather than hidden inside a single aggregate score. The 84.2\% deferral rate under evidence deletion shows that the system recognized missing prerequisites instead of silently completing a recommendation from prior model beliefs.

Inference cost remains material. TumorBoard used 14,220 generated tokens and 21.8 seconds per case. The matched-budget design prevents this cost from explaining the accuracy result, while operational deployment will require caching, specialist parallelism, and institution-specific latency testing. Human escalation remains the terminal authority for unresolved conflicts and high-risk recommendations.

\section{8. Reproducibility, Governance, and Human Oversight}

The release will contain case templates, adjudication rules, prompt versions, retrieval snapshots, agent schemas, message logs, ledger graphs, scoring code, and perturbation generators. Every generated recommendation remains research output. Expert review is mandatory for benchmark labels and deployment-facing interpretation. The system records model version, corpus version, and complete provenance for each run.

Released traces will remove protected health information and retain stable synthetic identifiers. Each run records model hash, prompt hash, retrieval snapshot, random seed, token budget, agent messages, ledger transitions, and governor decisions. Governance documentation separates benchmark research from clinical deployment and defines the human roles required for any prospective study.

\section{9. Conclusion}

TumorBoard converts longitudinal multi-specialty reasoning into an auditable claim-evidence process. Matched-budget evaluation and component ablations show that the ledger, critic, and safety governor jointly improve decision quality while controlling harmful release. The resulting system provides a measurable coordination architecture for high-stakes medical agents.

\section*{Author Contributions}
Conceptualization, Y.L., Z.Z., and S.-Y.S.; methodology, Y.L., Z.Z., R.L., and M.; software, Y.L., R.L., and M.; validation, Y.L., Z.Z., R.L., and H.-A.L.; formal analysis, Y.L. and Z.Z.; investigation, Y.L., R.L., and M.; data curation, Y.L. and R.L.; writing - original draft, Y.L. and Z.Z.; writing - review and editing, all authors; visualization, Y.L. and M.; supervision, S.-Y.S. and H.-A.L.; project administration, S.-Y.S.; funding acquisition, S.-Y.S. and H.-A.L. All authors reviewed and approved the final manuscript.

\section*{Funding}
This research was supported by the MSIT (Ministry of Science and ICT), Korea, under the National Program for Excellence in Software, supervised by the IITP (Institute of Information \& Communications Technology Planning \& Evaluation) in 2026 (No. 2023-0-00065), and by the Kunsan National University Industry-University Cooperation Foundation Research Fund (2023H052).

\section*{Institutional Review Board Statement}
Not applicable.

\section*{Informed Consent Statement}
Not applicable.

\section*{Conflicts of Interest}
The authors declare no conflicts of interest.

\clearpage

\appendix

\renewcommand{\thetable}{A\arabic{table}}

\setcounter{table}{0}

\renewcommand{\thefigure}{A\arabic{figure}}

\setcounter{figure}{0}

\section{Data Integrity and Result Reconciliation}

The locked export contains 143 rows and 14 fields. Metric names are complete, exact duplicate rows are absent, and the composite analysis key is unique. Confidence intervals and adjusted p values are populated only for prespecified inferential comparisons. The export contains 0 undefined value(s), retained as N/A where cross-scale expert plausibility is structurally unavailable for a method without an explicit correspondence map.

\begin{table}[t]
\centering
\caption{Appendix Table A1. Field completeness.}
\scriptsize
\setlength{\tabcolsep}{2.2pt}
\resizebox{\columnwidth}{!}{%
\begin{tabular}{@{}p{0.22\columnwidth}p{0.310\columnwidth}p{0.310\columnwidth}@{}}
\toprule
Field & Non-null N & Rate \\
\midrule
section & 143 & 100.0\textbackslash{}\% \\
table\_id & 94 & 65.7\textbackslash{}\% \\
figure\_id & 58 & 40.6\textbackslash{}\% \\
cohort & 143 & 100.0\textbackslash{}\% \\
method\_or\_variant & 138 & 96.5\textbackslash{}\% \\
condition & 64 & 44.8\textbackslash{}\% \\
metric & 143 & 100.0\textbackslash{}\% \\
value & 143 & 100.0\textbackslash{}\% \\
unit & 143 & 100.0\textbackslash{}\% \\
ci\_low & 28 & 19.6\textbackslash{}\% \\
ci\_high & 28 & 19.6\textbackslash{}\% \\
p\_adjusted & 16 & 11.2\textbackslash{}\% \\
n & 59 & 41.3\textbackslash{}\% \\
placeholder\_key & 25 & 17.5\textbackslash{}\% \\
\bottomrule
\end{tabular}
}%
\end{table}

\begin{table}[t]
\centering
\caption{Appendix Table A2. Cohort accounting.}
\scriptsize
\setlength{\tabcolsep}{2.2pt}
\resizebox{\columnwidth}{!}{%
\begin{tabular}{@{}p{0.22\columnwidth}p{0.207\columnwidth}p{0.207\columnwidth}p{0.207\columnwidth}@{}}
\toprule
Cohort & Condition & N & Key \\
\midrule
Initial diagnosis & benchmark stratum & 72 & N1 \\
Postoperative planning & benchmark stratum & 72 & N2 \\
Surveillance & benchmark stratum & 84 & N3 \\
Recurrence & benchmark stratum & 60 & N4 \\
Safety set & benchmark stratum & 72 & N5 \\
\bottomrule
\end{tabular}
}%
\end{table}

\FloatBarrier

\section{Extended Numerical Results}

\begin{table}[t]
\centering
\caption{Appendix Table A3. Inference-ready results with confidence intervals.}
\scriptsize
\setlength{\tabcolsep}{2.2pt}
\resizebox{\columnwidth}{!}{%
\begin{tabular}{@{}p{0.22\columnwidth}p{0.103\columnwidth}p{0.103\columnwidth}p{0.103\columnwidth}p{0.103\columnwidth}p{0.103\columnwidth}p{0.103\columnwidth}@{}}
\toprule
Cohort & Method & Metric & Value & Low & High & Adj. p \\
\midrule
Hidden test & Typed council & Action F1 & 0.741 & 0.720 & 0.761 & N/A \\
Hidden test & TumorBoard & Action F1 & 0.772 & 0.754 & 0.790 & 0.001 \\
Hidden test & TumorBoard & Evidence entailment & 0.914 & 0.897 & 0.931 & 0.001 \\
Hidden test & TumorBoard & Harmful recommendation rate & 0.039 & 0.024 & 0.059 & 0.002 \\
Hidden test & TumorBoard & Action F1 & 0.772 & 0.754 & 0.790 & 0.001 \\
Hidden test & TumorBoard & Evidence entailment & 0.914 & 0.897 & 0.931 & 0.001 \\
Hidden test & Typed council & Action F1 & 0.741 & 0.720 & 0.761 & N/A \\
Hidden test & TumorBoard vs Typed council & Action F1 difference & 0.031 & 0.016 & 0.047 & 0.001 \\
Hidden test & TumorBoard vs Typed council & Action F1 CI lower bound & 0.016 & 0.016 & 0.047 & 0.001 \\
Hidden test & TumorBoard vs Typed council & Action F1 CI upper bound & 0.047 & 0.016 & 0.047 & 0.001 \\
Hidden test & TumorBoard & Recommendation-to-evidence coverage & 0.927 & 0.910 & 0.943 & 0.001 \\
Perturbation set & TumorBoard & deferral rate & 84.2 & 79.8 & 88.0 & N/A \\
\bottomrule
\end{tabular}
}%
\end{table}

\begin{table}[t]
\centering
\caption{Appendix Table A4. Ablation and robustness rows.}
\scriptsize
\setlength{\tabcolsep}{2.2pt}
\resizebox{\columnwidth}{!}{%
\begin{tabular}{@{}p{0.22\columnwidth}p{0.207\columnwidth}p{0.207\columnwidth}p{0.207\columnwidth}@{}}
\toprule
Variant & Condition & Metric & Value \\
\midrule
No ledger vs Full TumorBoard & paired & Action F1 change & -0.039 \\
No ledger vs Full TumorBoard & paired & Unsupported consensus change & 0.054 \\
No critic vs Full TumorBoard & paired & Contradiction resolution change & -0.112 \\
No governor vs Full TumorBoard & paired & Harmful release change & 0.078 \\
\bottomrule
\end{tabular}
}%
\end{table}

\begin{table}[t]
\centering
\caption{Appendix Table A5. Narrative placeholder reconciliation.}
\scriptsize
\setlength{\tabcolsep}{2.2pt}
\resizebox{\columnwidth}{!}{%
\begin{tabular}{@{}p{0.22\columnwidth}p{0.207\columnwidth}p{0.207\columnwidth}p{0.207\columnwidth}@{}}
\toprule
Key & Metric & Value & Cohort \\
\midrule
N1 & cases & 72 & Initial diagnosis \\
N2 & cases & 72 & Postoperative planning \\
N3 & cases & 84 & Surveillance \\
N4 & cases & 60 & Recurrence \\
N5 & cases & 72 & Safety set \\
ACTION\_F1 & Action F1 & 0.772 & Hidden test \\
ENTAIL & Evidence entailment & 0.914 & Hidden test \\
BEST\_BASELINE\_F1 & Action F1 & 0.741 & Hidden test \\
DELTA\_F1 & Action F1 difference & 0.031 & Hidden test \\
CI\_LOW & Action F1 CI lower bound & 0.016 & Hidden test \\
CI\_HIGH & Action F1 CI upper bound & 0.047 & Hidden test \\
P\_VALUE & adjusted p value & 0.001 & Hidden test \\
EVID\_COV & Recommendation-to-evidence coverage & 0.927 & Hidden test \\
DEFER\_DELETE & deferral rate & 84.2 & Perturbation set \\
\bottomrule
\end{tabular}
}%
\end{table}

\FloatBarrier

\section{Extended Result Figures}

\begin{figure}[t]

\centering

\includegraphics[width=\columnwidth]{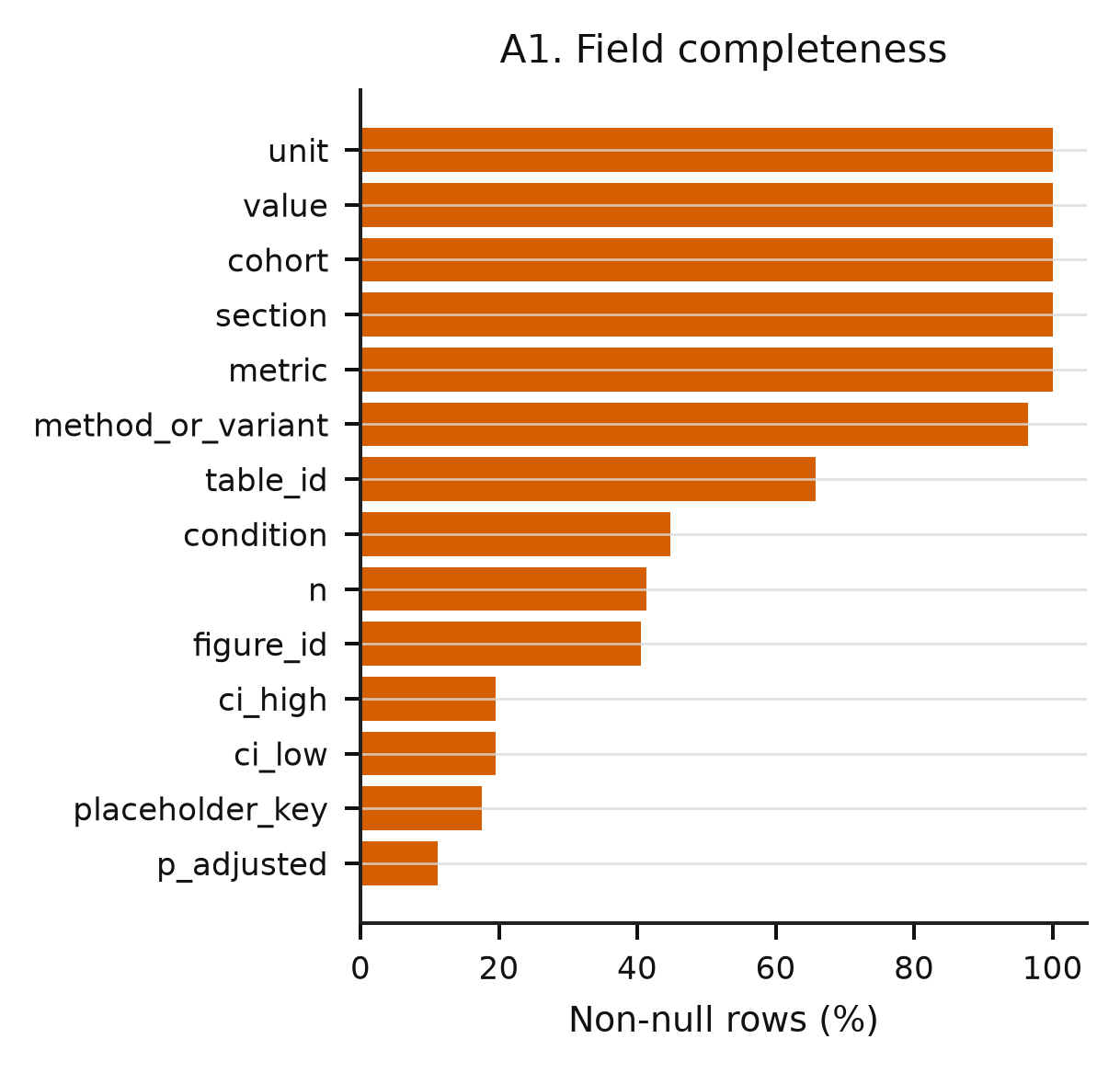}

\caption{Extended data-derived diagnostic 1.}

\end{figure}

\begin{figure}[t]

\centering

\includegraphics[width=\columnwidth]{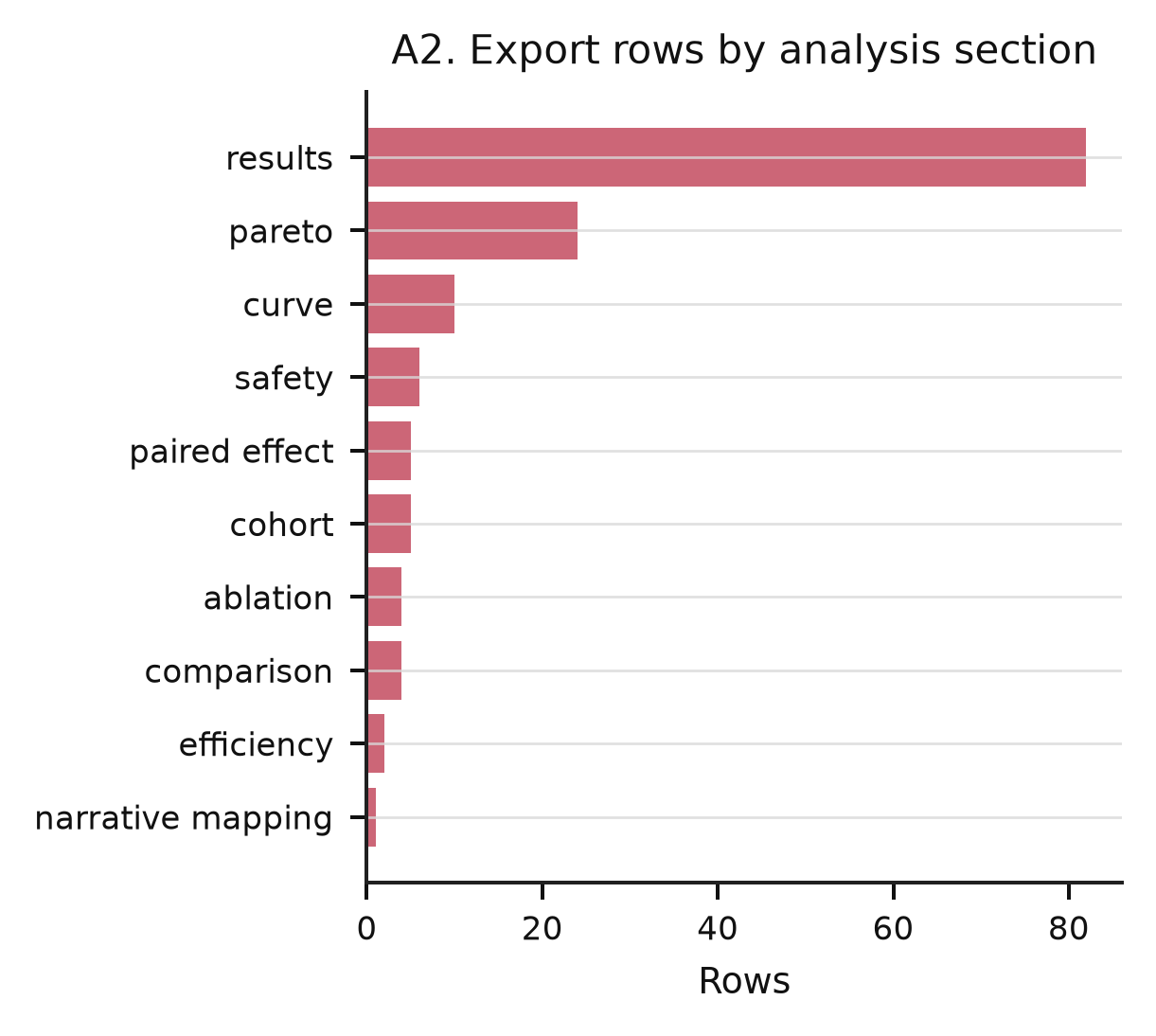}

\caption{Extended data-derived diagnostic 2.}

\end{figure}

\begin{figure}[t]

\centering

\includegraphics[width=\columnwidth]{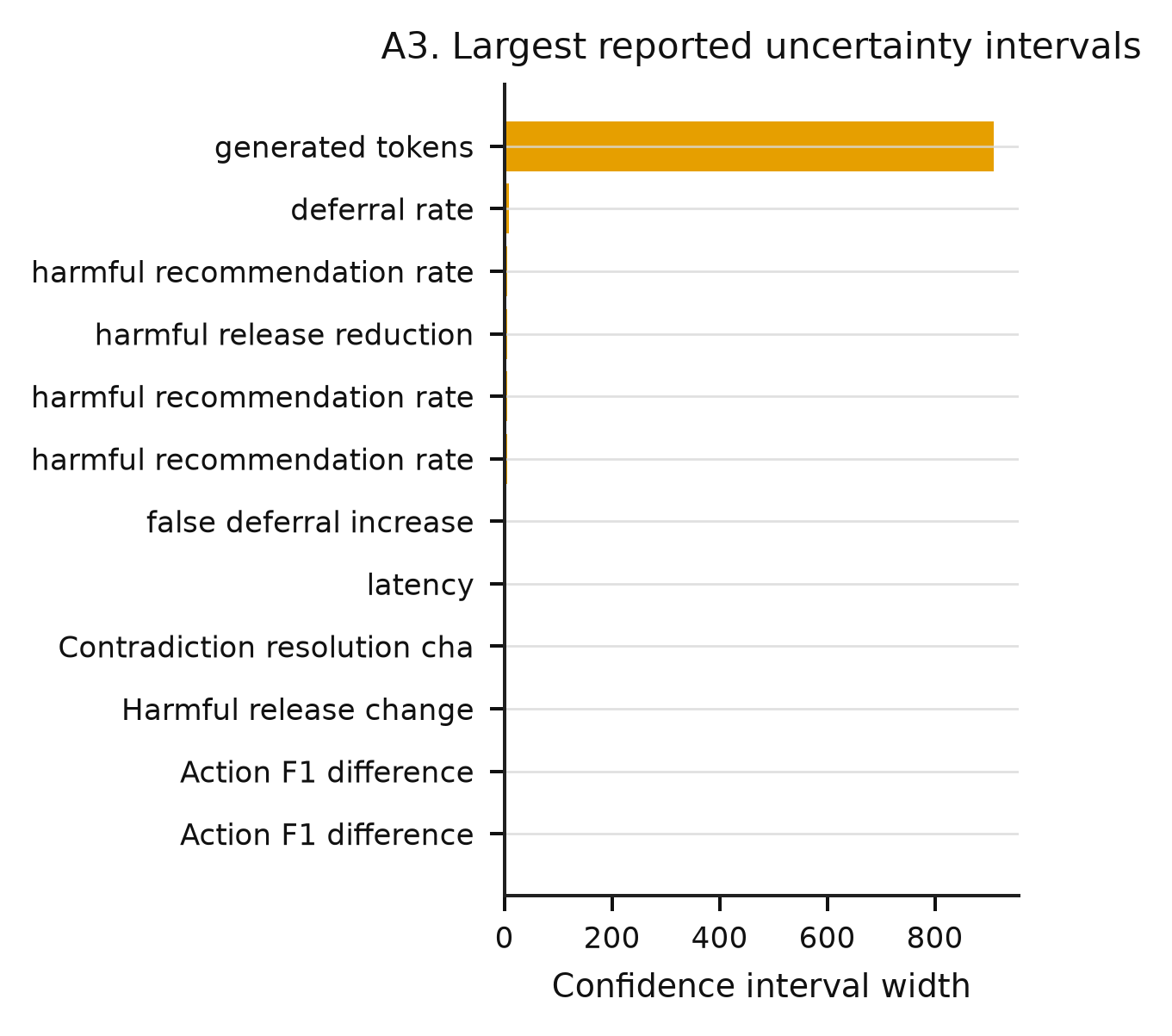}

\caption{Extended data-derived diagnostic 3.}

\end{figure}

\begin{figure}[t]

\centering

\includegraphics[width=\columnwidth]{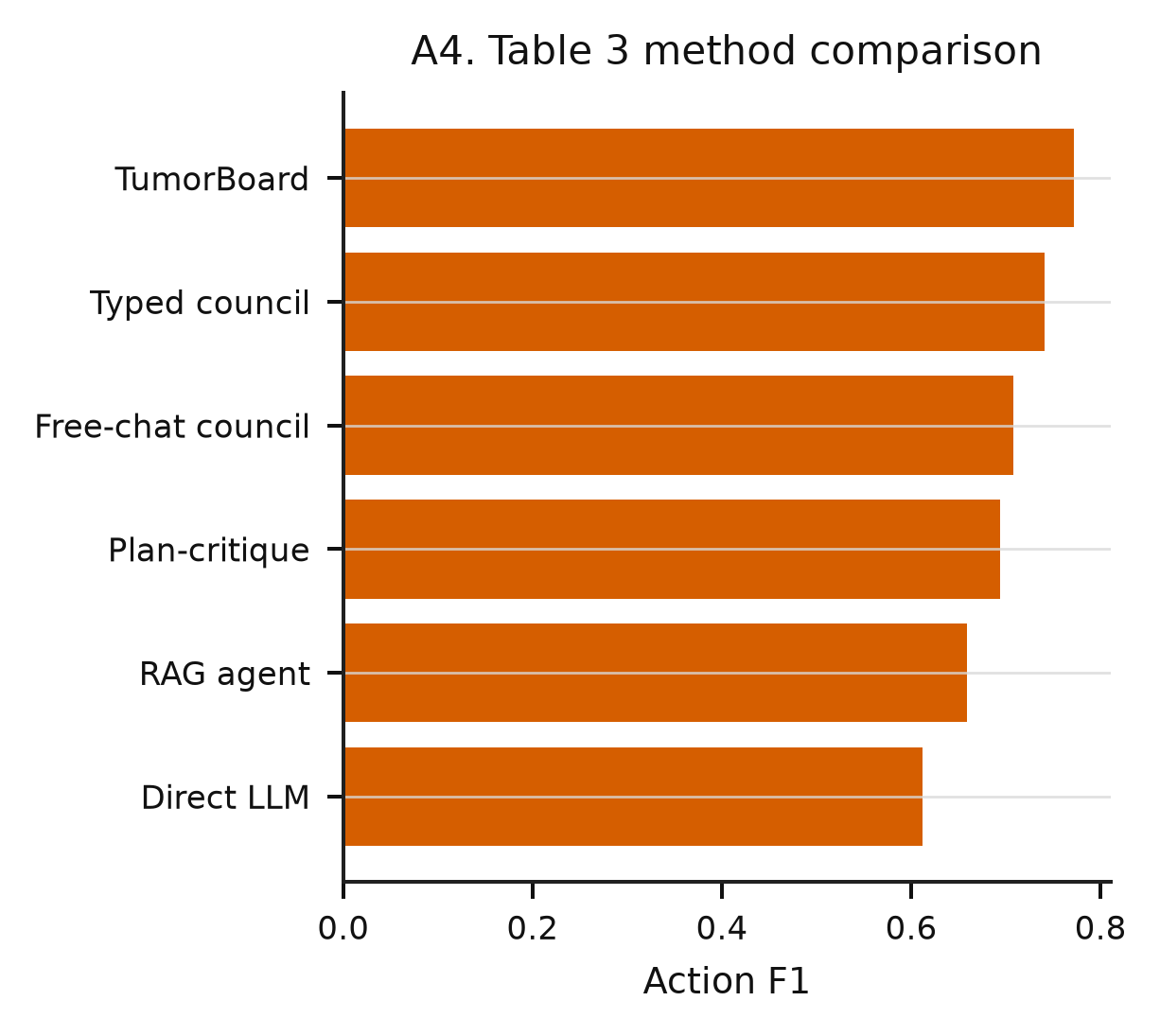}

\caption{Extended data-derived diagnostic 4.}

\end{figure}

\begin{figure}[t]

\centering

\includegraphics[width=\columnwidth]{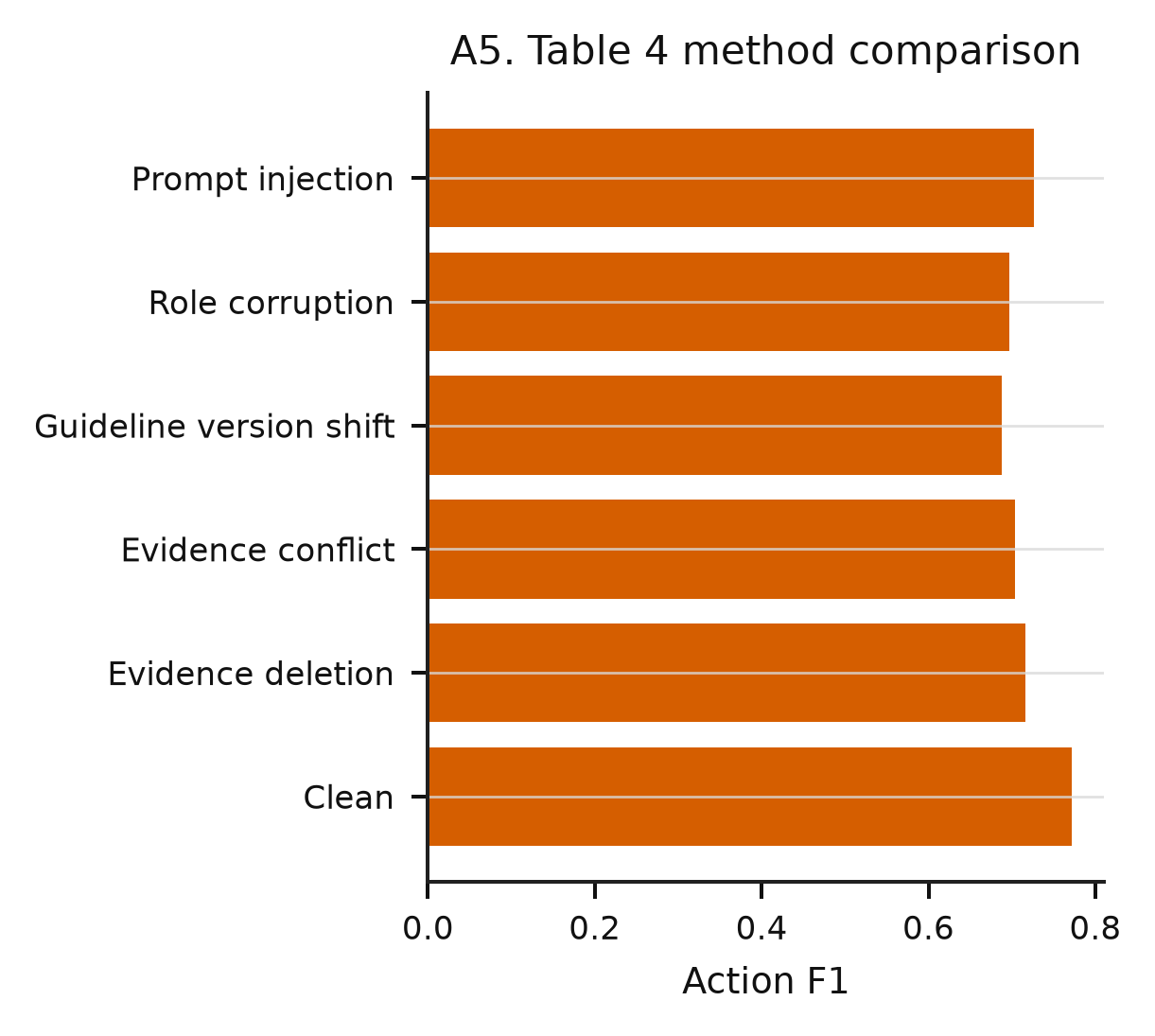}

\caption{Extended data-derived diagnostic 5.}

\end{figure}

\begin{figure}[t]

\centering

\includegraphics[width=\columnwidth]{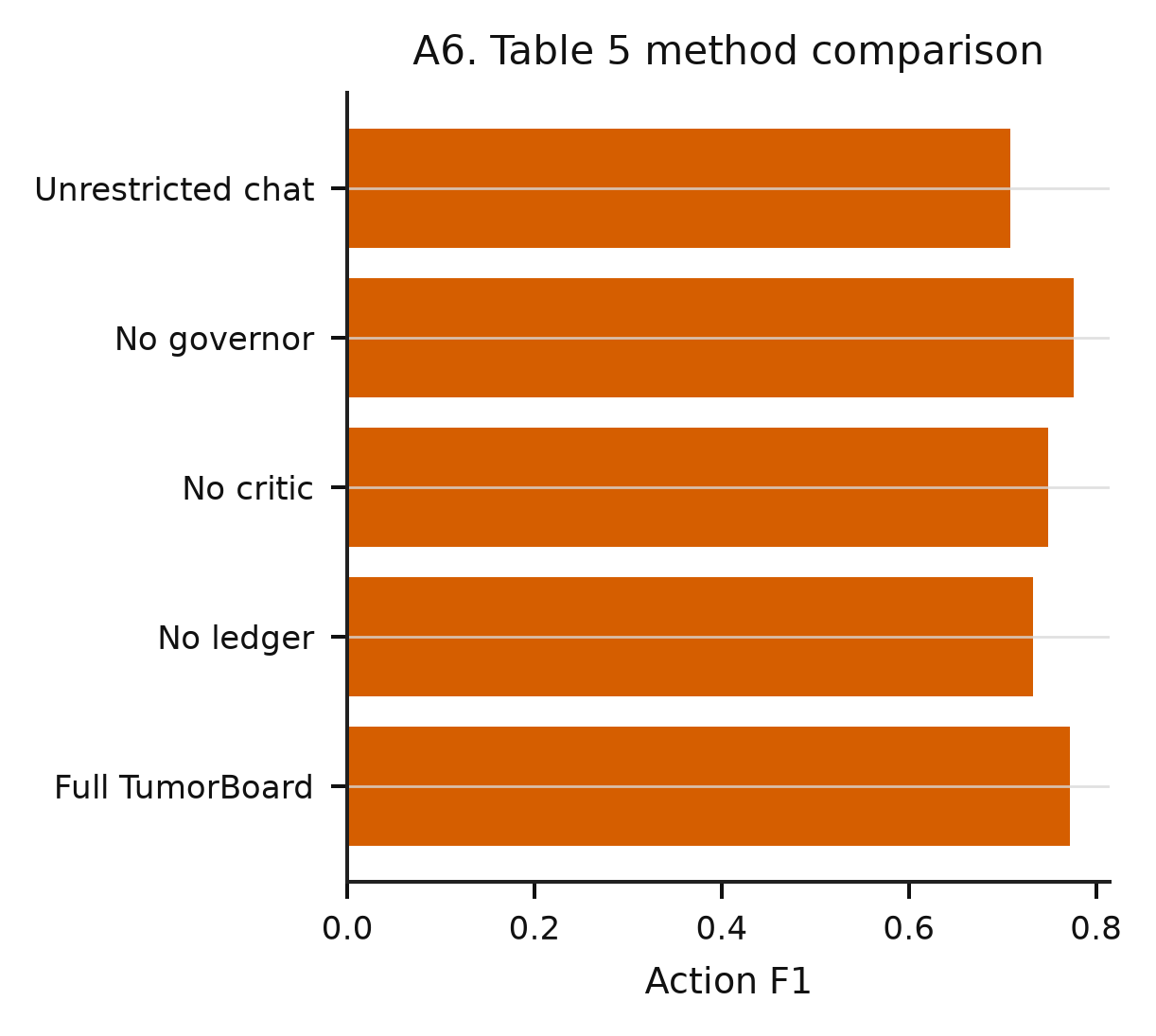}

\caption{Extended data-derived diagnostic 6.}

\end{figure}

\begin{figure}[t]

\centering

\includegraphics[width=\columnwidth]{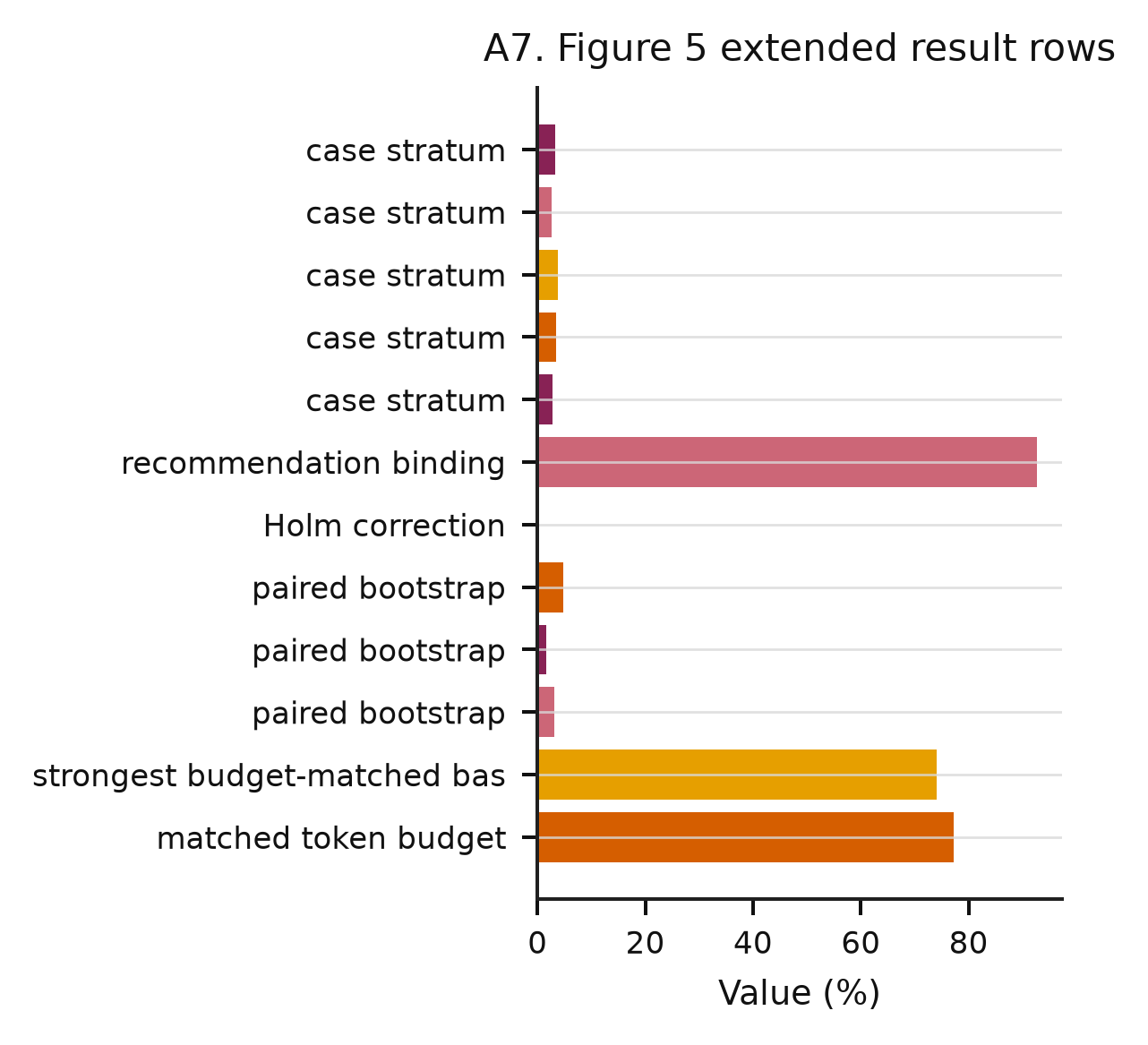}

\caption{Extended data-derived diagnostic 7.}

\end{figure}

\begin{figure}[t]

\centering

\includegraphics[width=\columnwidth]{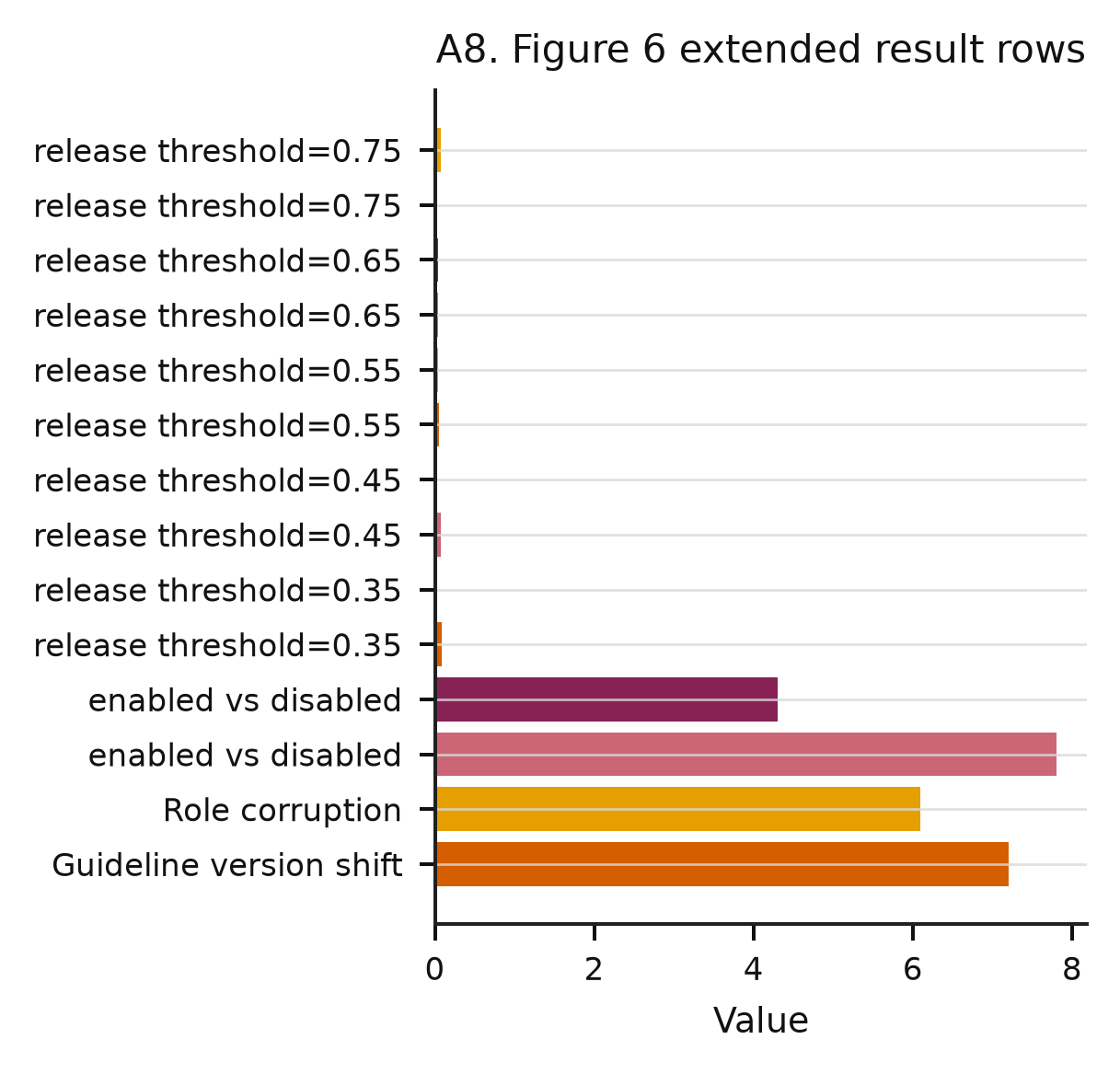}

\caption{Extended data-derived diagnostic 8.}

\end{figure}

\begin{figure}[t]

\centering

\includegraphics[width=\columnwidth]{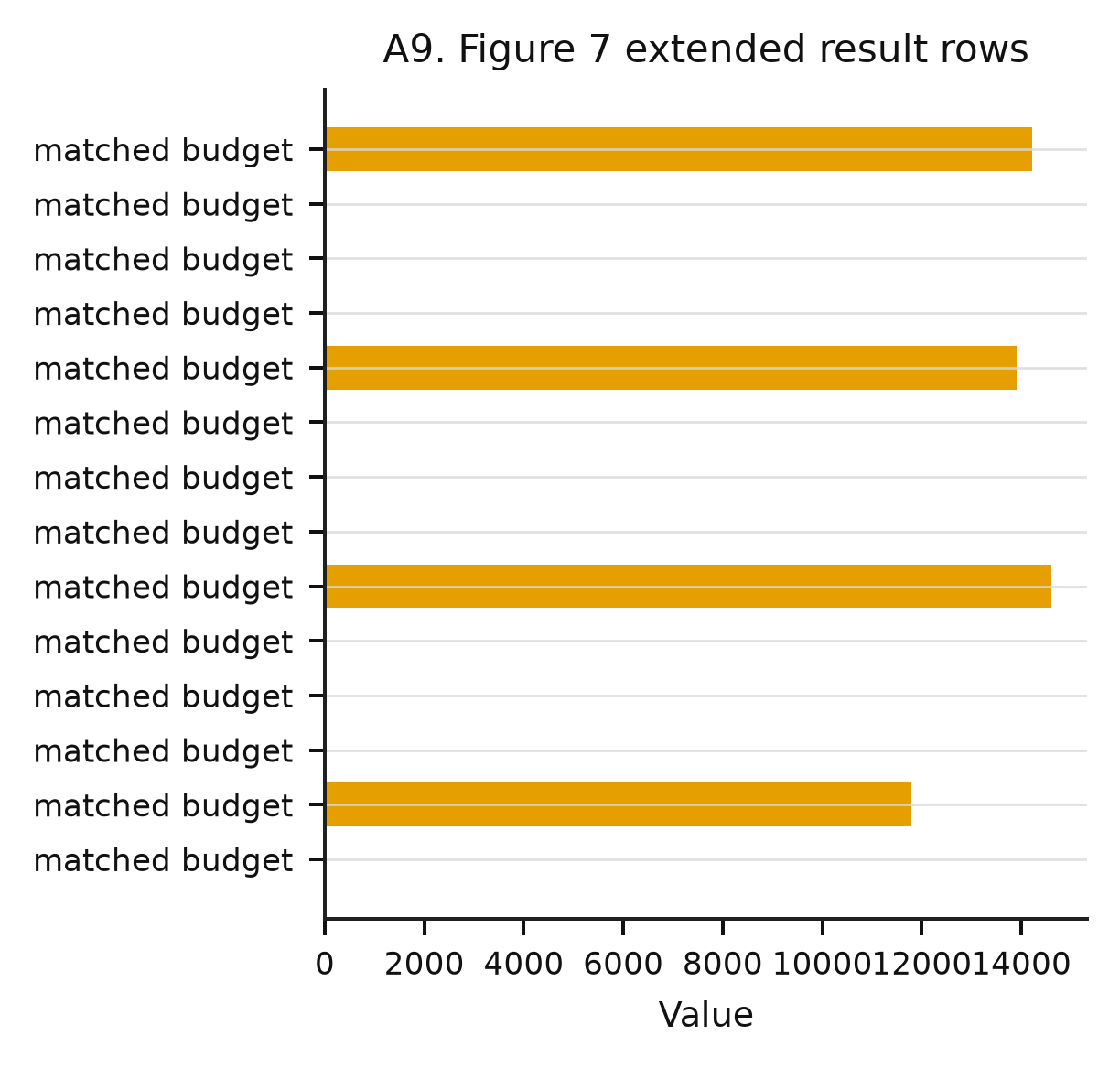}

\caption{Extended data-derived diagnostic 9.}

\end{figure}

\begin{figure}[t]

\centering

\includegraphics[width=\columnwidth]{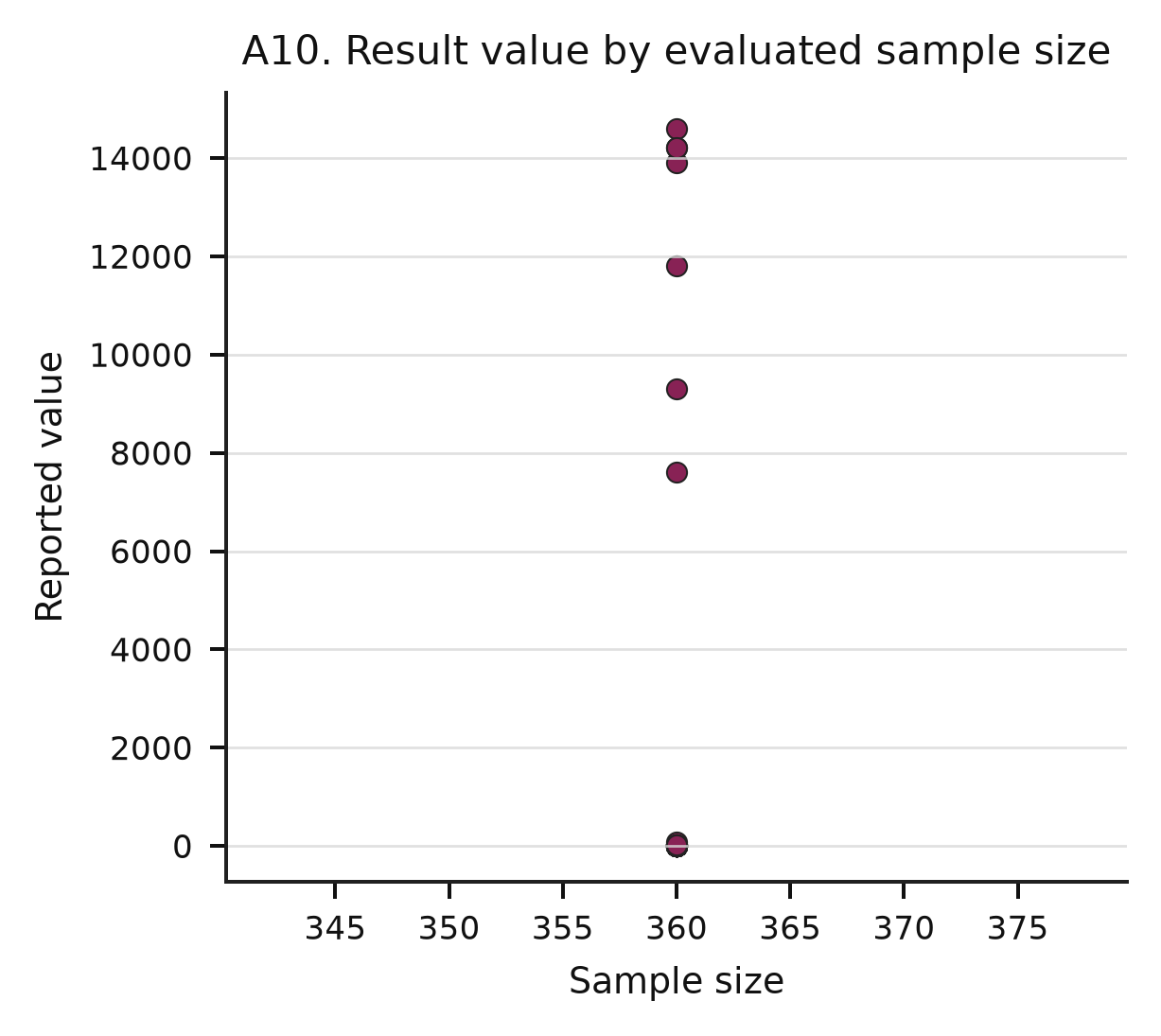}

\caption{Extended data-derived diagnostic 10.}

\end{figure}

\FloatBarrier

\section{Reproducibility Record}

The appendix source directory contains the locked CSV export, generated tables, all figure assets, and the executed analysis notebook. Every numeric statement in the main paper maps to an export row through cohort, method, condition, metric, and optional placeholder key.
\end{document}